\documentclass[conference]{IEEEtran}

\usepackage[utf8]{inputenc}
\usepackage{amssymb,amsfonts,amsmath,amscd}
\usepackage{bm}
\usepackage[pdftex]{graphicx}
\usepackage{url}
\usepackage{cite}
\usepackage[usenames,dvipsnames]{color}
\usepackage{subcaption}
\usepackage{caption}
\usepackage{accents}
\usepackage{tabularx}
\usepackage{tikz}
\usepackage{stackengine}
\usepackage{hyperref}

\usepackage{enumitem}

\usepackage{mathtools}
\usepackage[english]{babel}
\usepackage{siunitx}
\usepackage[export]{adjustbox}

\usepackage{microtype}

\usepackage{booktabs}
\usepackage{makecell}
\usepackage{multirow}

\definecolor{pltred}{rgb}{0.839, 0.153, 0.157}

\DeclareMathOperator*{\maximize}{maximize}

\newcommand{\R}{\mathbb{R}}

\newcommand{\Rot}{\bm{C}}
\newcommand{\stilde}{\skew5\tilde}

\newcommand{\videourl}{http://tiny.cc/visual-servor}
\newcommand{\codeurl}{https://github.com/learnsyslab/visual\_servor}

\title{Robotic Nonprehensile Object Transportation with a Hanging Tray}
\author{\IEEEauthorblockN{Adam Heins}
\IEEEauthorblockA{\textit{University of Toronto} \\
Toronto, Canada \\
adam.heins@robotics.utias.utoronto.ca}
\and
\IEEEauthorblockN{Angela P. Schoellig}
\IEEEauthorblockA{\textit{Technical University of Munich} \\
Munich, Germany \\
angela.schoellig@tum.de}
}

\begin{document}

\maketitle

\begin{abstract}
  We consider the nonprehensile object transportation task known as the
  \emph{waiter's problem}, in which a robot must move an object balanced on a
  tray from one location to another. In contrast to prior works on the robotic
  waiter's problem, which make the robot tilt a tray \emph{rigidly held} by its
  end effector (EE), we use a tray \emph{suspended} from the EE by
  ropes, such that it behaves like a three-dimensional pendulum. Some prior
  works have actuated the robot so that the EE \emph{simulates} the behavior of
  a pendulum, because pendular motion reduces the shear forces acting on the
  transported objects, minimizing the sliding of rigid objects and sloshing in
  containers of liquid. In contrast, our use of a \emph{real} hanging tray
  allows us to obtain the benefits of pendular motion while only actuating a
  3~degree-of-freedom (DOF) mobile base, rather than requiring a full 6-DOF
  manipulator arm. Our experiments in simulation and on real hardware show that
  the hanging tray substantially reduces both sliding and sloshing compared to
  a static, rigidly-grasped tray. Furthermore, we integrate the hanging tray
  into an interactive robot waiter demonstration, which uses computer vision to
  identify people with a raised hand and visual servoing to steer toward them
  and allow them to access the tray.
\end{abstract}

\section{Introduction}

The \emph{waiter's problem}~\cite{flores2013time} is a nonprehensile
manipulation task that requires a robot to transport objects balanced on a
tray, like a restaurant waiter. Nonprehensile
manipulation~\cite{lynch1996nonprehensile} occurs when the manipulated objects
are not rigidly grasped, such that they retain some independent degrees of
freedom (DOFs). In the waiter's problem, the objects are only attached to the
tray by frictional contact, and thus it is possible for them to slide or tip
unless the motion of the robot is designed to avoid it. Other examples of
nonprehensile manipulation include pushing, rolling, and
throwing~\cite{heins2024force,ruggiero2018nonprehensile}. Solving the waiter's
problem is appealing because the use of a tray avoids the need to grasp and
ungrasp each object, allows many objects to be transported at once, and can
handle delicate objects which may be difficult to grasp at
all~\cite{pham2017admissible}. It may be useful for transporting objects in
industries including food service, warehouse fulfillment, and manufacturing.

\begin{figure}[t]
  \centering
  \footnotesize
  \begin{tikzpicture}
    \draw (0, 0) node[inner sep=0] {\fbox{\includegraphics[width=\columnwidth]{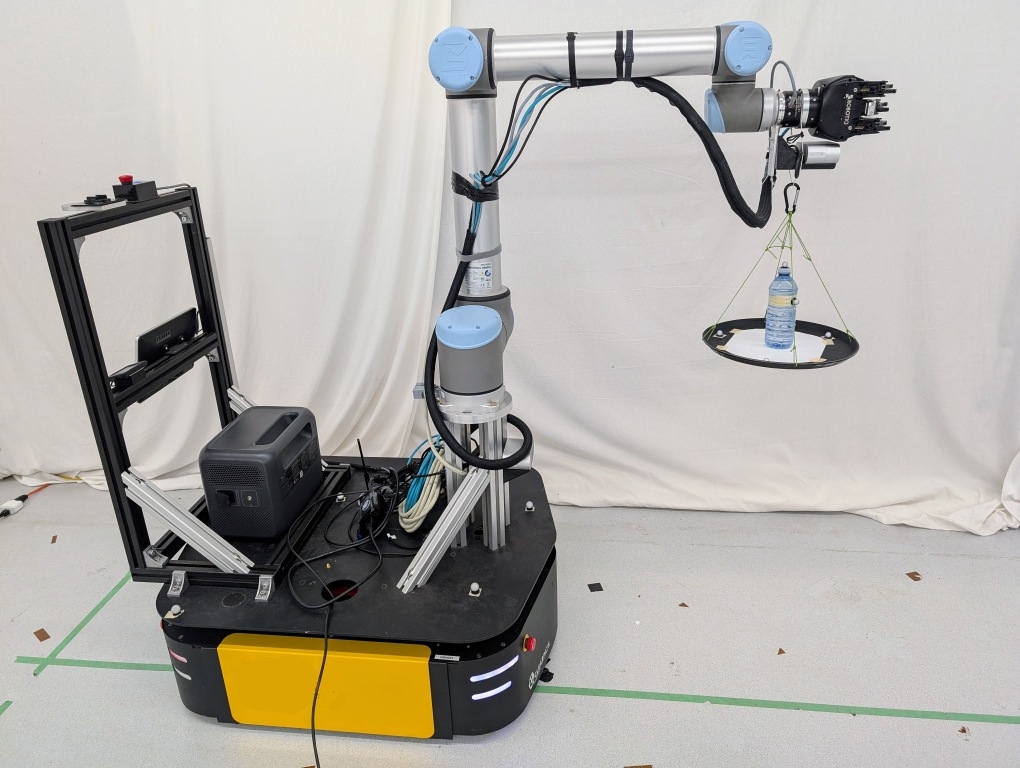}}};
    \draw[red,very thick] (2.3, 0.8) ellipse (1cm and 1cm);
    \draw[blue,very thick] (2.6, 1.95) ellipse (0.4cm and 0.4cm);
  \end{tikzpicture}
  \caption{We present the first work on robotic nonprehensile object
    transportation that uses a real hanging tray (circled red) to support the
  transported objects. A key benefit is that an actuated manipulator arm is not
  necessary: in all of our experiments, we keep the arm stationary and only
  actuate the mobile base,
  allowing the swinging of the tray to naturally prevent the objects from sliding.
  An RGB-D camera (circled blue) allows the robot to detect human gestures and
  steer toward people to serve them. A video of the experiments can be found at
  \textbf{\texttt{\scriptsize \videourl}}.}
  \label{fig:eyecandy}
\end{figure}

In contrast to previous approaches to the robotic waiter's problem, which tilt
a tray \emph{rigidly attached} to the robot's end effector (EE), we borrow an
idea from human waiters (e.g., the Morrocan tea service~\cite{dang2004active})
and \emph{suspend} the tray from ropes, allowing it to swing freely like a
three-dimensional (3D) pendulum. The swinging naturally reduces the shear
forces acting on the transported objects, therefore requiring less friction to
prevent the objects from sliding. A key benefit of this approach is that the
robot can move quickly without needing to rotate the EE to prevent sliding.
Indeed, in our experiments on the robot shown in Fig.~\ref{fig:eyecandy}, we
only control the mobile base and do not require any actuation of the
manipulator arm. Once the robot reaches its destination, the swinging of the
tray is rapidly damped out using a linear quadratic regulator (LQR).

In addition, we integrate the hanging tray into an interactive robot waiter
demonstration that uses an RGB-D camera to detect and serve people like a
waiter. In particular, we use a convolutional neural network (CNN) to detect
people raising their hands, indicating that they would like to be served. The
robot then uses visual servoing to steer toward them. When the robot reaches
the person, it damps out the swinging of the tray, waits a fixed amount of time
for the person to take the desired item from the tray, and then returns to its
original position until another person gestures to it. This is the first time
that gesture-based visual servoing has been used explicitly for the waiter's
problem.

In summary, the main contributions of this work are:
\begin{itemize}
  \item the first use of a real hanging tray for robotic nonprehensile object
    transportation, including modelling, analysis, and hardware experiments;
  \item a novel interactive robot waiter demonstration that combines the
    hanging tray with gesture recognition and visual servoing to interactively
    serve people, demonstrated on a real mobile manipulator;
  \item an open-source implementation of our code, available at
    \textbf{\texttt{\footnotesize\codeurl}}.
\end{itemize}

\section{Related Work}

The waiter's problem has been approached using a variety of methods including
offline
planning~\cite{pham2017admissible,zhou2022topp,gattringer2023point,brei2024serving},
online planning (i.e., model predictive
control)~\cite{selvaggio2023non,heins2023keep}, and reactive
control~\cite{moriello2018manipulating,muchacho2022a,selvaggio2022a,subburaman2023a}.
Broadly, all of these methods try to tilt the tray so that the shear forces
acting on the transported object are small, and therefore friction can balance
them out and prevent the object from sliding. However, this at least
requires the actuated DOFs to tilt the tray, and may also require an object
model that includes geometry and inertial parameters.

One interesting approach is to simulate the motion of a pendulum with the
robot's EE, which naturally minimizes shear forces acting on the transported
object without explicitly modelling it. This was first done
in~\cite{dang2004active} with a parallel manipulator mounted on a mobile robot,
but only in two dimensions. This idea was extended to the three-dimensional
case~\cite{moriello2018manipulating,muchacho2022a} to minimize slosh when
transporting liquids with a fixed-base manipulator arm. Inspired by these
approaches, we use a real, physical pendulum rather than simulating one. That
is, we \emph{suspend} the tray from the EE so that it can swing freely, which
means we do not need an actuated arm to tilt the tray nor do we need models
of the transported objects. Our proposed approach is similar to the commercial
SpillNot product~\cite{millstein2012hanging}, which is a suspended tray
designed for carrying cups of liquid without spilling them. However, we are the
first to use a suspended tray for nonprehensile object transportation on a
robot, which presents the additional challenge of damping out the unwanted
swinging of the tray once the robot reaches its destination. To do so, we
design an LQR-based controller similar to~\cite{august20103d}, except that we
also introduce an integral term to eliminate the steady-state position error
that occurs when the pendulum model is imperfect. We opt for LQR over
anti-swing planning methods like~\cite{chen2007swing} to avoid the need to
follow a planned trajectory, which would limit reactivity and increase
complexity.

Furthermore, we integrate the hanging tray into an interactive robot waiter
demonstration, inspired by previous robot waiters like
Alfred~\cite{maxwell1999alfred}. We use a CNN to detect people with
a raised hand and use visual servoing to steer toward them. Visual
servoing---that is, closed-loop motion control based on visual input---has a
long history~\cite{hutchinson1996a}, and gesture recognition for human-robot
interaction has also been extensively
studied~\cite{liu2018gesture,qi2024computer}, but fewer works have combined
combined the two. In~\cite{waldherr2000a}, a mobile robot uses
visual servoing to track and follow a person while also identifying gestures
made by the person to trigger a behavior change, such as to start or stop a
task. Similarly, \cite{sawadwuthikul2022visual} trains a classifier to
identify a particular person in a camera feed given a reference image of that person,
which allows a mobile robot to dynamically track an individual around the
environment and deliver items to them, but does not use gesture recognition.
In~\cite{alshanoon2018vision}, a CNN is trained to identify hand gestures. A
fixed-based manipulator then uses visual servoing to track the gesturing hand
with its EE. Likewise, in~\cite{zhang2024a}, a custom neural network is
trained to recognize hand gestures, which are used to command both a
manipulator and a mobile robot with manipulation targets and movement commands,
respectively. However, our work is the first time gesture-based visual servoing
has been used explicitly for the waiter's problem; that is, visual servoing is
used to steer a mobile robot toward a gesturing person while bringing them an
object transported on a tray.

\section{System Models}

Here we develop models of the 3D pendulum and combined tray-object system,
which we use for subsequent control design and numerical experiments. We denote
the~$n\times n$ identity matrix as~$\bm{1}_n$, the set of~$n\times n$ symmetric
positive semidefinite matrices as~$\mathbb{S}^n_+$, and the set of~$n\times n$
symmetric positive definite matrices as~$\mathbb{S}^n_{++}$.

\subsection{Pendulum Model}

The dynamics of a rigid body can be expressed using the body-frame Newton-Euler
equations:
\begin{equation*}
  \bm{w} = \bm{M}\dot{\bm{\xi}} - \mathrm{ad}(\bm{\xi})^T\bm{M}\bm{\xi},
\end{equation*}
where~$\bm{w}=[\bm{\tau}^T,\bm{f}^T]^T$ is the applied wrench with
torque~$\bm{\tau}\in\R^3$ and force~$\bm{f}\in\R^3$, $\bm{M}\in\mathbb{S}^6_+$
is the body's spatial mass matrix, $\bm{\xi}=[\bm{\omega}^T,\bm{v}^T]^T$ is the
spatial velocity with angular component~$\bm{\omega}\in\R^3$ and linear
component~$\bm{v}\in\R^3$, and
\begin{equation*}
  \mathrm{ad}(\bm{\xi}) \triangleq \begin{bmatrix}
    \bm{\omega}^\times & \bm{0} \\ \bm{v}^\times & \bm{\omega}^{\times}
  \end{bmatrix}
\end{equation*}
is the adjoint of~$\bm{\xi}$ with~$(\cdot)^{\times}$ forming a skew-symmetric
matrix such that~$\bm{a}^{\times}\bm{b}=\bm{a}\times\bm{b}$ for
any~$\bm{a},\bm{b}\in\R^3$. The spatial mass matrix is defined as
\begin{equation*}
  \bm{M} = \begin{bmatrix} \bm{I} & \bm{h}^\times \\ -\bm{h}^\times & m\bm{1}_3 \end{bmatrix},
\end{equation*}
where~$\bm{I}\in\mathbb{S}^3_+$ is the inertia matrix and~$\bm{h}=m\bm{c}$ is
the first moment of mass with mass~$m>0$ and center of mass (CoM)~$\bm{c}\in\R^3$.

A 3D pendulum~\cite{shen2004dynamics} is a rigid body that is constrained to
rotate about a pivot point~$O$; that is, its position is fixed but it retains
all three rotational degrees of freedom. Taking~$O$ as the origin of the body
frame and assuming the only external force is gravity, the dynamics simplify to
Euler's equation:
\begin{equation}\label{eq:eulers_eq}
  \bm{h}^\times\Rot^T\bm{g} = \bm{I}\dot{\bm{\omega}}+\bm{\omega}^\times\bm{I}\bm{\omega},
\end{equation}
where~$\Rot\in SO(3)$ is the pendulum's orientation and $\bm{g}\in\R^3$ is the
gravitational acceleration expressed in the global inertial frame. The
orientation~$\Rot$ and body-frame angular velocity~$\bm{\omega}$ are related
by the kinematics equation~$\dot{\Rot}=\Rot\bm{\omega}^{\times}$.

In our case, the pivot point is attached to the robot's EE, which is itself
moving. Let~$\bm{u}\in\R^3$ be the acceleration of the pivot point in the
global frame. Then the dynamics~\eqref{eq:eulers_eq} are augmented to
\begin{equation}\label{eq:3d_pendulum}
  \bm{h}^\times\Rot^T(\bm{g}-\bm{u}) = \bm{I}\dot{\bm{\omega}}+\bm{\omega}^\times\bm{I}\bm{\omega}.
\end{equation}

If we assume~$\bm{c}^T\bm{\omega}=0$ and that all mass is
concentrated at~$\bm{c}$ (such that~$\bm{I}=-m\bm{c}^{\times}\bm{c}^{\times}$),
then the system is known as a spherical or 2D pendulum, and the
dynamics~\eqref{eq:3d_pendulum} simplify to
\begin{equation}\label{eq:2d_pendulum}
  \bm{c}^\times\Rot^T(\bm{g}-\bm{u}) = \ell^2\dot{\bm{\omega}},
\end{equation}
where~$\ell=\|\bm{c}\|_2$.

\subsection{Tray-Object Model}

\begin{figure}[t]
  \vspace{5pt}
  \centering
  \includegraphics[width=0.8\columnwidth]{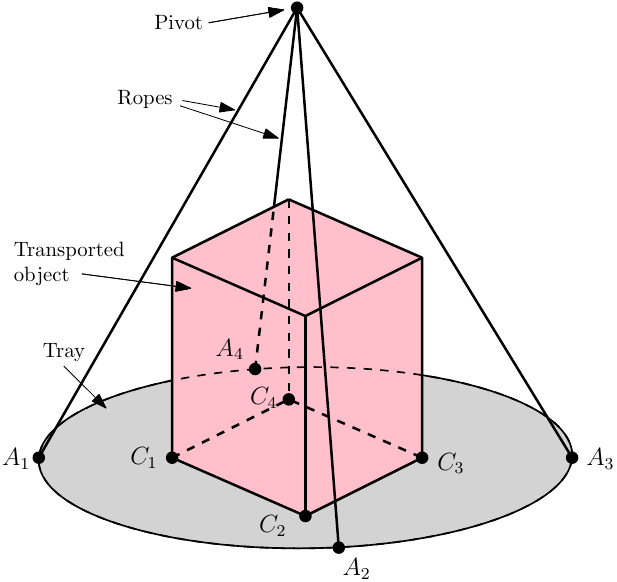}
  \caption{A hanging tray (gray) supporting an object (red). The tray is
  suspended from a pivot point by ropes attached to the anchor points
  $A_1$--$A_4$. We model contact between the object and tray using a discrete
  set of contact points $C_1$--$C_4$ at the vertices of the contact area.}
  \label{fig:3d_diagram}
\end{figure}

Suppose that the hanging tray is supporting an object, which we assume is a rigid
body, as shown in Fig.~\ref{fig:3d_diagram}. Let~$\bm{M}_o$ be the object's
spatial mass matrix and let~$\bm{M}_t$ be the tray's spatial mass matrix. Let
us assume that there is no relative motion between the object and tray, so that
they share the same orientation~$\Rot$ and spatial velocity~$\bm{\xi}$. Given a
commanded EE acceleration~$\bm{u}\in\R^3$, we can solve for their shared
spatial acceleration using~\eqref{eq:3d_pendulum} and the kinematic
relationship~$\bm{u}=d(\bm{C}\bm{v})/dt$ to obtain
\begin{equation*}
  \dot{\bm{\xi}} = \begin{bmatrix} \dot{\bm{\omega}} \\ \dot{\bm{v}} \end{bmatrix}
  = \begin{bmatrix}
    \stilde{\bm{I}}^{-1}(\tilde{\bm{h}}^\times\Rot^T(\bm{g}-\bm{u}) - \bm{\omega}^\times\stilde{\bm{I}}\bm{\omega}) \\
    \Rot^T\bm{u} - \bm{\omega}^\times\bm{v}
  \end{bmatrix},
\end{equation*}
where
\begin{equation*}
  \tilde{\bm{M}} = \begin{bmatrix}
    \stilde{\bm{I}} & \tilde{\bm{h}}^\times \\ -\tilde{\bm{h}}^\times & \tilde{m}\bm{1}_3
  \end{bmatrix} = \bm{M}_t + \bm{M}_o
\end{equation*}
is the spatial mass matrix of the combined system.

We assume that the object and tray interact through a set of~$n_c$ contact
forces~$\{\bm{f}_{c_i}\}_{i=1}^{n_c}$, located at
points~$\{\bm{r}_{c_i}\}_{i=1}^{n_c}$. To ensure that our assumption that there is
no relative motion between the two bodies holds, we must be able to find a set
of feasible contact forces; that is, the contact forces must lie within their
Coulomb friction cones, defined by
\begin{equation}\label{eq:fc}
  \sigma_{c_i}(\bm{f}_{c_i}) \triangleq \mu_{c_i}\hat{\bm{n}}_{c_i}^T\bm{f}_{c_i} - \|\hat{\bm{n}}_{c_i}^\times\hat{\bm{n}}_{c_i}^\times\bm{f}_{c_i}\|_2 \geq 0,
\end{equation}
where~$\mu_{c_i}\geq0$ is the friction coefficient
and~$\hat{\bm{n}}_{c_i}\in\R^3$ is the unit-length contact normal. In our
numerical experiments, it will be useful to relax~\eqref{eq:fc} to
\begin{equation}\label{eq:fc_relaxed}
  \sigma(\bm{f}_{c_i}) \geq s
\end{equation}
with slack variable~$s\in\R$, where~$s\geq0$ indicates that the original
constraint~\eqref{eq:fc} is feasible.

Furthermore, we suspend the tray from the pivot point on the EE using~$n_a$
ropes attached to the tray at anchor points~$\{\bm{r}_{a_i}\}_{i=1}^{n_a}$, so we
also need to constrain the tensile forces~$f_{a_i}\in\R$, $i=1,\dots,n_a$,
along each rope to be non-negative. That is, we need to find contact and
tensile forces that satisfy the following system of constraints:
\begin{equation}\label{eq:force_constraints}
  \begin{aligned}
    \bm{w}_a - \bm{w}_c + m_t\bm{G}(\bm{c}_t)\Rot^T\bm{g} &= \bm{M}_{t}\dot{\bm{\xi}} - \mathrm{ad}(\bm{\xi})^T\bm{M}_{t}\bm{\xi}, \\
    \bm{w}_c + m_o\bm{G}(\bm{c}_o)\Rot^T\bm{g} &= \bm{M}_{o}\dot{\bm{\xi}} - \mathrm{ad}(\bm{\xi})^T\bm{M}_{o}\bm{\xi}, \\
    \sigma(\bm{f}_{c_i}) &\geq s, \quad i=1,\dots,n_c, \\
    f_{a_i} &\geq 0, \quad i=1,\dots,n_a,
  \end{aligned}
\end{equation}
where the first two lines are the Newton-Euler equations for the tray and
object, respectively, with tensile and contact wrenches
\begin{align*}
  \bm{w}_a &= \sum_i^{n_a}f_{a_i}\bm{G}(\bm{r}_{a_i})\bm{r}_{a_i}, & \bm{w}_c &= \sum_i^{n_c}\bm{G}(\bm{r}_{c_i})\bm{f}_{c_i},
\end{align*}
where~$\bm{G}(\bm{p}) = [-\bm{p}^{\times}, \bm{1}_3]^T$
for any~$\bm{p}\in\R^3$. To facilitate comparison in our numerical experiments,
we will determine the feasible contact forces by solving the following convex
optimization problem:\footnote{Using the friction cone constraint~\eqref{eq:fc}
or its relaxed version~\eqref{eq:fc_relaxed} results in a second-order cone
program. It is also common to linearize~\eqref{eq:fc},
making~\eqref{eq:force_constraints_relaxed} a linear program.}
\begin{equation}\label{eq:force_constraints_relaxed}
  \begin{aligned}
    s^\star = \maximize & \quad s \\
    \text{subject to} & \quad \eqref{eq:force_constraints},
  \end{aligned}
\end{equation}
which finds the contact forces with the largest friction constraint margin.
If~$s^\star\geq0$, then all of the contact forces are feasible under the Coulomb
model~\eqref{eq:fc}. In contrast, $s^\star<0$ implies that no feasible contact
forces can be found to resist the object's motion, at which point we stop the
simulation. The controller itself does not depend on the contact forces, but
they must be computed while \emph{simulating} the system to determine if the
object moves relative to the tray.

\section{Swing Attenuation}\label{sec:lqr}

After the robot's EE reaches a desired location and stops, the tray will
continue to swing until all of its energy is lost to friction and damping. We
would like to speed up this process and rapidly bring the tray to rest, to
facilitate a person interacting with objects on it. To do so, we design a
linear-quadratic regulator (LQR) using the simplified 2D pendulum
model~\eqref{eq:2d_pendulum} linearized about the hanging equilibrium point,
which is a similar approach to~\cite{august20103d} except that we also introduce an
integral term to eliminate steady-state pivot position error. It is convenient
to parameterize~\eqref{eq:2d_pendulum} with a unit vector~$\bm{\delta}\in\R^3$
pointing from the pivot position~$\bm{r}\in\R^3$ toward the pendulum's CoM,
rather than the rotation matrix~$\bm{C}$. The state
is~$\bm{x}=[\bm{\gamma}^T,\bm{r}^T,\bm{\delta}^T,\dot{\bm{r}}^T,\dot{\bm{\delta}}^T]^T\in\R^{15}$, where~$\bm{\gamma}\in\R^3$ is the time integral of~$\bm{r}$ (that
is, $\bm{\gamma}$ is the pivot's \emph{absement}), with all quantities expressed in the
global frame. We found that the addition of an integral term is necessary to
eliminate steady-state error in the presence of model errors like an inaccurate
pendulum CoM, as we will demonstrate in Sec.~\ref{sec:sim}. The nonlinear
equations of motion are
\begin{equation*}
  \dot{\bm{x}} = \bm{f}(\bm{x},\bm{u}) = \begin{bmatrix}
    \bm{r} \\ \dot{\bm{r}} \\ \dot{\bm{\delta}} \\ \bm{u} \\
    -\|\dot{\bm{\delta}}\|^2\bm{\delta} - \bm{\delta}^\times\bm{\delta}^\times(\bm{g}-\bm{u})/\ell
  \end{bmatrix},
\end{equation*}
which we linearize by computing the
Jacobians~$\bm{A}\triangleq\partial\bm{f}/\partial\bm{x}$
and~$\bm{B}\triangleq\partial\bm{f}/\partial\bm{u}$ about the hanging
equilibrium. However, the $z$-component of~$\bm{\delta}$ is not controllable in
this linearized system and our mobile robot cannot change the $z$-component
of~$\bm{r}$. We therefore use the reduced state~$\bm{x}'\in\R^{10}$ and
input~$\bm{u}'\in\R^2$, which drop all $z$-components, and corresponding
reduced matrices~$\bm{A}'\in\R^{10\times10}$ and~$\bm{B}'\in\R^{10\times2}$.
This gives us the linearized
system~$\dot{\bm{x}}'=\bm{A}'\bm{x}'+\bm{B}'\bm{u}'$, for which we use LQR to
obtain the linear feedback law
\begin{equation}\label{eq:lqr_ctrl}
  \bm{u}'=-\bm{K}(\bm{x}'-\bm{x}_d'),
\end{equation}
where~$\bm{x}_d=[\bm{\gamma}_d^T,\bm{r}_d^T,\bm{0}^T,\bm{0}^T,\bm{0}^T]^T$
with~$\bm{r}_d$ the desired pivot position and~$\bm{\gamma}_d$ its integral.
The gain matrix is~$\bm{K}=\bm{R}^{-1}\bm{B}'^T\bm{P}$, where~$\bm{P}$ is obtained by
solving the continuous-time algebraic Ricatti equation
\begin{equation*}
  \bm{A}'^T\bm{P}+\bm{P}\bm{A}' - \bm{P}\bm{B}'\bm{R}^{-1}\bm{B}'^T\bm{P} + \bm{Q} = \bm{0}
\end{equation*}
given tunable gain matrices~$\bm{Q}\in\mathbb{S}^{10}_{++}$
and~$\bm{R}\in\mathbb{S}^2_{++}$. In all experiments we
use~$\bm{Q}=\bm{1}_{10}$ and~$\bm{R}=0.1\bm{1}_2$.

\section{Interactive Robot Waiter}\label{sec:interactive}

We now describe how to integrate the hanging tray into an interactive robot
waiter system, where the use of a hanging tray is appealing because it allows
the robot to move freely without worrying about the transported objects
falling. An RGB-D camera is used to observe the scene and detect humans to
serve.

\subsection{State Machine}

\begin{figure}[t]
  \vspace{2pt}
  \centering
  \includegraphics[width=0.75\columnwidth]{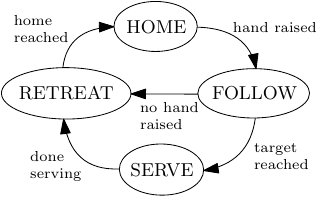}
  \caption{State machine for our interactive robot waiter demonstration with four
    states. The robot waits at the HOME position until a person raises their
  hand. It then FOLLOWs them using visual servoing until it arrives and SERVEs them
  or the hand is lowered; in either case, the robot then RETREATs back to the
  home position until another hand is raised.}
  \label{fig:state_machine}
  \vspace{-2pt}
\end{figure}

The robot's behavior is dictated by a simple state machine with four states,
depicted in Fig.~\ref{fig:state_machine}. They are:
\begin{itemize}
  \item HOME: the robot starts at a known home position in the room. If a
    person within view of the camera raises their hand, the robot starts to
    FOLLOW the person.
  \item FOLLOW: the robot moves forward while using visual servoing to steer
    toward the person with their hand up. If the person puts their hand down
    and there are no other people with hands up, then the robot RETREATs. If
    the robot reaches the target person, then it SERVEs them.
  \item SERVE: once the robot is within a threshold distance to the target
    person (as measured by the depth camera), the robot stops, attenuates the
    swinging of the tray using the control law~\eqref{eq:lqr_ctrl}, and then
    waits a pre-defined amount of time for the person to interact with the
    tray. Once done, the robot RETREATs.
  \item RETREAT: the robot moves back to the home position. If a person with
    their hand up is observed, the robot switches to FOLLOWing them.
\end{itemize}

Acceleration limits are enforced whenever the velocity changes between states.
The hand gesture detection and visual servoing procedures are described in more
detail below.

\subsection{Gesture Recognition and Visual Servoing}

The scene is observed using an RGB-D camera mounted at the EE. We use a
customized YOLOv11 model~\cite{yolo11_ultralytics} to obtain segmentation masks
of people with a hand raised. We use the COCO-Pose
dataset~\cite{lin2014microsoft}, which provides labelled human segmentation
masks and pose keypoints. The keypoints define a skeleton of 17 points,
including the nose, each eye, each ear, and each wrist. We process the dataset
so that each human segmentation mask is labelled either \emph{hand-up} or
\emph{hand-down}, using the following procedure. For each human instance in
each image, we compute the head height~$h_{\mathrm{head}}$ (in image
coordinates) as the mean value of the height of the nose, eye, and ear keypoints,
omitting any of these keypoints that are not visible. The instance is
classified as \emph{hand-up} if either wrist keypoint is
above~$h_{\mathrm{head}}$, and \emph{hand-down} otherwise. We then re-train the
YOLOv11 segmentation model on this dataset to predict (only) these two classes
of segmentation mask. The upshot is that we obtain a model that robustly
segments people with a raised hand from images.

Given an image containing a person with a raised hand and that person's
segmentation mask, we compute a horizontal target~$\alpha_x$ for the robot to
steer toward using visual servoing.\footnote{Since we are only controlling the
mobile base, we only care about a horizontal target. If the arm were being
used, it could raise the EE up and down to also align with a vertical target.}
In particular, we take~$\alpha_x$ to be the median of the horizontal values of the
segmentation mask, normalized to the interval~$[-1,1]$ with the origin at the center
of the image.
The robot moves forward with linear velocity~$v=\bar{v}(1-|\alpha_x|)$ and
angular velocity~$\omega=k_{\omega}\alpha_x$, where~$\bar{v},k_\omega>0$ are
tunable parameters. The velocity~$v$ moves the robot toward the target but is
reduced as~$|\alpha_x|$ increases, while~$\omega$ rotates the robot to
reduce~$|\alpha_x|$ (i.e., to center the target in the camera's view). We
compute the median depth over the segmentation mask to determine the current
distance to the target person.

\section{Numerical Experiments}\label{sec:sim}

We now present numerical simulations to provide insight into how the hanging tray
and object system described by~\eqref{eq:force_constraints}
behaves. We simulate the system forward in time from initial
state~$(\bm{r}_0,\bm{C}_0,\bm{\xi}_0)=(\bm{0},\bm{1}_3,\bm{0})$. Given the
pivot acceleration~$\bm{u}_k$ at each simulation timestep~$k$, we
solve~\eqref{eq:force_constraints_relaxed} to look for feasible forces and
then use the semi-implicit Lie-Euler method to integrate the system to
timestep~$k+1$:
\begin{align*}
  \dot{\bm{r}}_{k+1} &= \dot{\bm{r}}_k + \Delta t\bm{u}, &
  \bm{\omega}_{k+1} &= \bm{\omega}_k + \Delta t\dot{\bm{\omega}}_{k+1}, \\
  \bm{r}_{k+1} &= \bm{r}_k + \Delta t\dot{\bm{r}}_{k+1}, &
  \Rot_{k+1} &= \Rot_k\exp(\Delta t\bm{\omega}^{\times}_{k+1}),
\end{align*}
where~$\exp(\cdot)$ is the matrix exponential and~$\Delta t=\SI{10}{ms}$ is the
integration timestep. This approach ensures that~$\Rot$ stays on the~$SO(3)$
manifold. We use the Clarabel solver~\cite{goulart2024clarabel} via
CVXPY~\cite{diamond2016cvxpy} to solve~\eqref{eq:force_constraints_relaxed}.

\subsection{Robustness to Parameter Variation}

In the theoretical limiting case when both the tray and object are point masses
located at the same position, no shear forces are applied to the object no
matter the pivot acceleration, and no sliding occurs. While not physically
realizable, this case agrees with the intuition that an object with its mass
concentrated closer to the tray is less likely to move. Here we explore how
different parameters of the system affect its ability to resist movement of the
object with respect to the tray.

\begin{figure}[t]
  \vspace{3pt}
  \centering
  \includegraphics[width=\columnwidth]{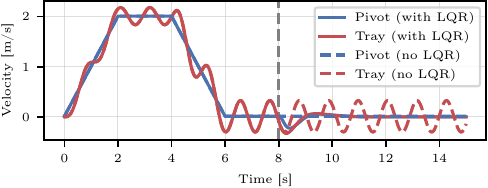}
  \caption{Simulated pivot and tray velocities, with and without LQR-based
  swing attenuation. When the LQR controller is activated at~$t=\SI{8}{s}$
  (dashed vertical line), it quickly damps out the tray's velocity by briefly
  moving the pivot. Without it, the tray simply keeps swinging.}
  \label{fig:sim_stab_vs_nostab}
\end{figure}

We model the tray as a thin disk with radius~\SI{20}{cm} and
thickness~\SI{1}{mm} transporting a box with side
lengths~$\SI{10}{cm}\times\SI{10}{cm}\times h$ placed directly over the tray's
center point, where~$h>0$ is the height of the box, similar to the setup in
Fig.~\ref{fig:3d_diagram}. Both the tray and object have mass~\SI{0.5}{kg} and
a uniform mass distribution. We use~$n_c=4$ contact points between the tray and
object, located at the contact area's vertices, with friction
coefficient~$\mu=0.1$ (i.e., fairly slippery; indeed, lower and thus more
difficult than the measured values used in the real-world experiments below).
The tray is suspended from~$n_a=4$
evenly spaced ropes such that it hangs at a length~$\ell$ below the pivot
point. Gravitational acceleration is~$\bm{g}=[0,0,-9.81]^T\si{m/s\squared}$. We
apply a horizontal acceleration corresponding to a trapezoidal velocity profile
to the pivot point. As shown in Fig.~\ref{fig:sim_stab_vs_nostab}, the
trajectory consists of a constant acceleration~$a$ for~$\SI{2}{s}$, a constant
velocity for an additional~\SI{2}{s}, and another constant acceleration of~$-a$
for a final~\SI{2}{s}.

\begin{figure}[t]
  \centering
  \includegraphics[width=\columnwidth]{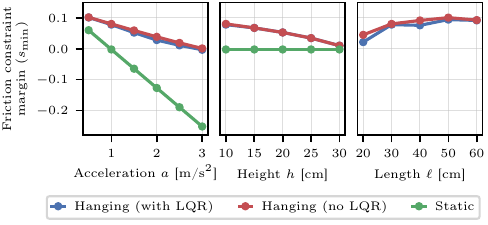}
  \caption{Friction constraint margin, defined as the minimum value~$s_{\min}$
    of~$s^\star$ obtained along the trajectory by
    solving~\eqref{eq:force_constraints_relaxed} at each simulation timestep.
    When~$s^\star<0$, no feasible set of friction forces exists, and the
    constraint is violated. The hanging tray results in higher constraint
    margins across a range of pivot accelerations and object heights compared to
    the static tray. Increasing the length of the pendulum also typically increases
    the constraint margin.}
  \label{fig:sim_param_variation}
\end{figure}

We vary the values of~$a$, $h$ and~$\ell$ and compare the friction constraint
satisfaction between a hanging tray and a ``static'' tray that is rigidly
attached to the pivot point (i.e., the EE).
For the hanging tray, we also compare the effect of swing attenuation using the
LQR control law~\eqref{eq:lqr_ctrl}. When LQR is used, we wait~\SI{2}{s} after
the end of the trajectory and then apply~\eqref{eq:lqr_ctrl} for~\SI{8}{s}. We
limit the magnitude of the pivot's acceleration input computed
from~\eqref{eq:lqr_ctrl} to no more than~\SI{1}{m/s\squared}. For each
parameter of interest, we vary it while the others are fixed at the nominal
values of~$a=\SI{1}{m/s\squared}$, $h=\SI{10}{cm}$, and~$\ell=\SI{30}{cm}$. As
can be seen from the results shown in Fig.~\ref{fig:sim_param_variation}, the
hanging tray can handle much higher accelerations without violating the
friction constraints. The static tray results in constraint violation for any
acceleration~$a>\mu\|\bm{g}\|=\SI{0.981}{m/s\squared}$, but the hanging tray
only starts to violate the constraints at~$a=\SI{3}{m/s\squared}$. The hanging
tray is also robust to different object heights~$h$: the constraint margin
decreases as height increases, but is still positive at~$h=\SI{30}{cm}$ with
the nominal acceleration~$a=\SI{1}{m/s\squared}$. In contrast, the constraint
violation with the static tray is independent of object height, and even
at~$a=\SI{1}{m/s\squared}$ the constraints are violated. Finally, we see that
increasing the length~$\ell$ of the pendulum typically increases the constraint
margin, while swing attenuation has a minimal impact on it.

\subsection{Swing Attenuation}

Let us now explore the behavior of swing attenuation in more detail.
Fig.~\ref{fig:sim_stab_vs_nostab} shows the velocities of the hanging tray and
pivot point using the nominal parameter values of~$a=\SI{1}{m/s\squared}$,
$h=\SI{10}{cm}$, and~$\ell=\SI{30}{cm}$, with and without LQR-based swing
attenuation. The trajectories are the same
until~$t=\SI{8}{s}$, when the LQR-based controller~\eqref{eq:lqr_ctrl} is
activated. With swing attenuation, the pivot moves to quickly damp out the
tray's velocity. Without it, the tray simply keeps swinging.

\begin{figure}[t]
  \centering
  \includegraphics[width=\columnwidth]{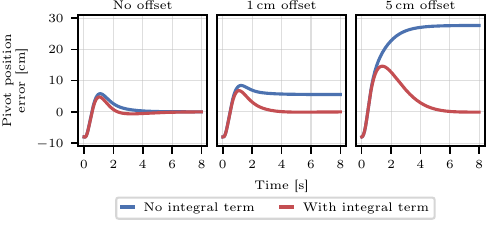}
  \caption{Effect of including the integral term~$\bm{\gamma}$ in the LQR-based
  swing attenuation controller on the pivot position error. When the
  transported object is offset from the center of the tray, such that the CoM of
  the combined system is no longer directly between the pivot and tray CoM,
  then substantial steady-state error occurs unless~$\bm{\gamma}$ is used.}
  \label{fig:sim_param_error}
\end{figure}

Next, we examine the effect of the integral term~$\bm{\gamma}$ we included in
the LQR model. The model assumes that the CoM of the combined tray-object
system is located along the line between the pivot point and the tray's CoM.
However, this assumption is easily violated if the object is placed such that
its CoM is horizontally offset from that of the tray. This offset results in
steady-state error in the pivot position unless~$\bm{\gamma}$ is included in
the LQR-based controller. We apply the same trapezoidal trajectory to the pivot
as in the previous subsection (again, using the nominal parameter values) while
varying the $x$-offset of the object with respect to the tray and we compare the
results with and without the integral term; the results are shown in
Fig.~\ref{fig:sim_param_error}. When there is no offset,
including~$\bm{\gamma}$ makes little difference. However, we see substantial steady-state
error in the pivot position when there is an offset unless~$\bm{\gamma}$ is
used. Indeed, with an offset of~\SI{5}{cm}, the resulting steady-state error
without~$\bm{\gamma}$ is over~\SI{25}{cm}; that is, the pivot ends up
over~\SI{25}{cm} from where we want it to be, which is not acceptable when
positioning the tray in a particular location to allow humans to interact with
it.

\subsection{A Failure Case}

Finally, let us consider an interesting failure mode of the hanging tray-object
system, which stems from the fact that it stores energy. This means that it is
possible to design an input pivot acceleration signal that pumps energy into
the system until the transported object slides and falls off the tray. Notably,
this can be achieved while keeping the EE acceleration below what is required
to make the object slide on a static tray. Let~$\dot{\bm{r}}_t$ be the velocity
of the tray's CoM in the global frame. Suppose the tray is moving, and we apply
the control law
\begin{equation}\label{eq:pump_energy}
  \bm{u} = \begin{cases}
    -a(\dot{\bm{r}}_t-\dot{\bm{r}})/\|\dot{\bm{r}}_t-\dot{\bm{r}}\|_2 & \text{if } \|\dot{\bm{r}}_t-\dot{\bm{r}}\|_2>0, \\
    \bm{0} & \text{otherwise},
  \end{cases}
\end{equation}
which applies an acceleration~$a$ in the opposite direction of the tray's
velocity relative to the pivot point. This is similar to the control law
designed in~\cite{aastrom2000swinging} to swing up a pendulum.

To demonstrate this failure case, we perform a simple simulation where the pivot point
accelerates horizontally at~$a=\SI{0.5}{m/s\squared}$ for~\SI{2}{s},
then switches to the control law~\eqref{eq:pump_energy}. With~$\mu=0.1$, the
acceleration is too low for any sliding to occur with a static tray. However,
the hanging tray system eventually fails to find feasible contact forces
at~$t=\SI{4.69}{s}$, indicating that the transported object would start to move
with respect to the tray. This failure mode is adversarial in nature, because
one has to apply a fairly specific input signal to achieve it. However, it
demonstrates how the hanging tray system is more nuanced than a static tray,
which only requires hard acceleration limits to prevent slipping.

\section{Hardware Experiments}

Having developed some intuition with numerical experiments, we now validate our
proposed approach with experiments on the real mobile manipulator shown in
Fig.~\ref{fig:eyecandy}, which consists of a 3-DOF Ridgeback omnidirectional
mobile base and a 6-DOF UR10 manipulator. However, since one of the key
motivations for using a hanging tray is that the EE does not need to be
actuated to prevent objects from sliding, we do not move the UR10 arm at all
in any experiments; all motion is achieved with the mobile base alone. The base
and hanging tray are localized using a Vicon motion capture system;
the tray's velocity is obtained using numerical differentiation. The robot is
velocity-controlled, so the acceleration input~\eqref{eq:lqr_ctrl} for
attenuating the tray's swinging is integrated in time to obtain velocity
commands. All experiments are run on a standard laptop with \SI{16}{GB} of RAM
and a NVIDIA Quadro M2200 GPU. We run our control loop at~\SI{25}{Hz}, which is
the frequency at which the base accepts commands. The reader is encouraged to
watch the video of the experiments at \textbf{\texttt{\small \videourl}}.

\subsection{Pendular Transport}

Let us compare the behavior of the hanging tray and static tray in the real
world. The static tray is held rigidly by the robot's gripper and the hanging
tray that is suspended from the EE using ropes, as shown in
Fig.~\ref{fig:tray_objects}.

\subsubsection{Rigid Objects}

\begin{table}[t]
  \vspace{10pt}
  \caption{Measured friction coefficients between objects and trays.}
  \centering
  \begin{tabular}{l c c}
    \toprule
    \multirow{2.5}*{Tray} & \multicolumn{2}{c}{Object} \\
    \cmidrule(lr){2-3}
    & Cup & Bottle \\
    \midrule
    Static & 0.34 & 0.22 \\
    \midrule
    Hanging & 0.26 & 0.19 \\
    \bottomrule
  \end{tabular}
  \label{tab:friction_coefficients}
\end{table}

\begin{figure}[t]
  \centering
  \footnotesize
  \fbox{\includegraphics[height=0.85in,trim={0 0 0 0},clip,valign=t]{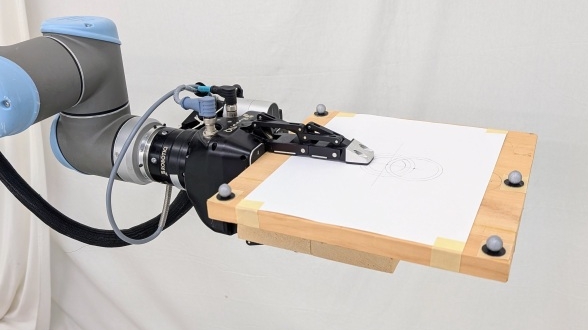}}
  \fbox{\includegraphics[height=0.85in,trim={0 0 0 0},clip,valign=t]{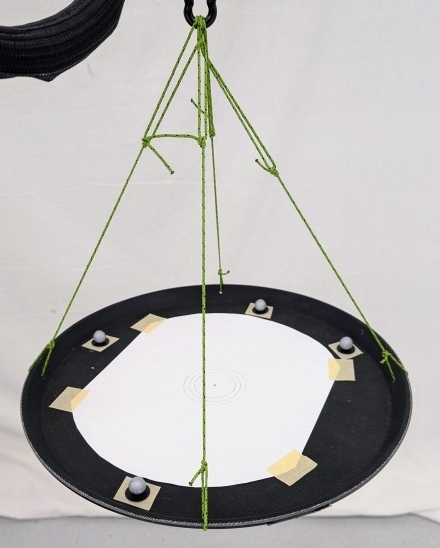}}
  \fbox{\includegraphics[height=0.85in,trim={0 0 0 0},clip,valign=t]{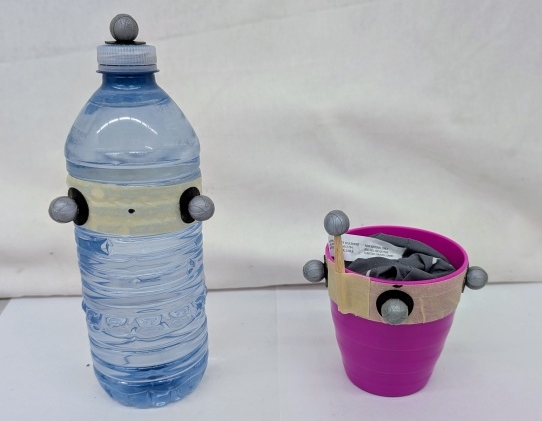}}
  \caption{\emph{From left:} The static tray, the hanging tray, and
  the two objects used in experiments: a water bottle and a cup.}
  \label{fig:tray_objects}
\end{figure}

We begin with experiments transporting two different (approximately) rigid
objects: a short cup and a taller water bottle, also shown in
Fig.~\ref{fig:tray_objects}. The cup contains two bean bags (to simulate being
filled without risking a spill) and the bottle is filled with water (but is
securely sealed). The cup is~\SI{8}{cm} tall and weighs~\SI{239}{g}. The bottle
is~\SI{19}{cm} tall and weighs~\SI{514}{g}. The two trays are made of different
materials, but we cover each with paper to provide similar surfaces. Despite
this, the measured friction coefficients (given in
Table~\ref{tab:friction_coefficients}) between the objects and the hanging tray
are slightly lower than with the static tray. However, this makes relative
motion \emph{more} likely with the proposed hanging tray, making
the comparison to the baseline static tray even stronger.

\begin{figure}[t]
  \centering
  \includegraphics[width=\columnwidth]{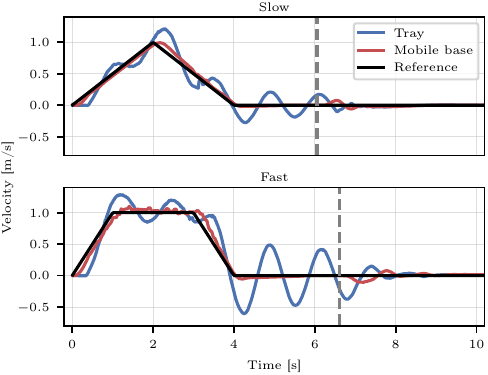}
  \caption{Velocity response of the mobile base and hanging tray for the slow and fast
  motion profiles. The vertical dashed line is the time at which swing
  attenuation is started, damping out the tray's velocity within a few seconds.
  The velocities are obtained by numerically differentiating the measured
  positions from the motion capture system and applying a small moving
  average filter.}
  \label{fig:velocity_response}
\end{figure}

To test the ability of each tray to transport each object, we command the base
to move in one direction with a particular motion profile and measure the
distance the object moves from its initial position on the tray using motion capture. Two
trapezoidal motion profiles are used: \emph{slow} and \emph{fast}. Starting
from rest, the \emph{slow} trajectory has a constant acceleration
of~\SI{0.5}{m/s\squared} for~$\SI{2}{s}$, reaching a maximum velocity
of~\SI{1}{m/s}, before immediately accelerating at~$-\SI{0.5}{m/s\squared}$ for
a further~\SI{2}{s} and returning to rest at~$t=\SI{4}{s}$. The \emph{fast}
trajectory accelerates from rest at~\SI{1}{m/s\squared} for~$\SI{1}{s}$ to
reach the same maximum velocity of~\SI{1}{m/s}. It holds this velocity constant
for~\SI{2}{s} before accelerating at~$-\SI{1}{m/s\squared}$ for a
further~\SI{1}{s} and returning to rest at~$t=\SI{4}{s}$. Both motion profiles
are shown in Fig.~\ref{fig:velocity_response}, along with the actual velocities
of the base and hanging tray from one of the experiments. When using the
hanging tray, we wait~\SI{1}{s} after the robot has come back to rest before
using the LQR-based controller~\eqref{eq:lqr_ctrl} to attenuate the swinging of
the tray; Fig.~\ref{fig:velocity_response} shows how the oscillations of the
tray are quickly damped out by briefly moving the mobile base.

\begin{figure}[t]
  \centering
  \includegraphics[width=\columnwidth]{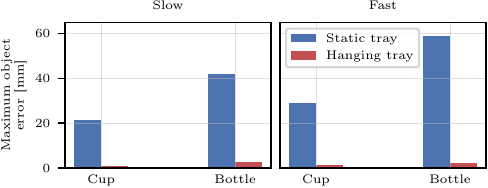}
  \caption{Maximum object error across different combinations of trajectory
  profile, object, and tray type. The error is the maximum distance that the object
  moved from its initial position relative to the tray; each value is the
  average across three trials, with 24 experiments in total. The hanging tray
  results in substantially lower error than the static tray, because its swinging
  naturally reduces the shear forces acting on the objects.}
  \label{fig:object_error}
\end{figure}

For each combination of transported object (cup or bottle), tray (static or
hanging), and motion profile (slow or fast), we perform three runs, for a total
of 24 experiments. The maximum object error across of any of the three runs for
each combination is plotted in Fig.~\ref{fig:object_error}. The hanging tray
resulted in almost no sliding of either object for both motion profiles, with
at most~\SI{3}{mm} of error. In contrast, the static tray results in
over~\SI{20}{mm} of error in every case. While the commanded accelerations
should theoretically be low enough to avoid sliding on the static tray,
unmodelled effects and vibrations result in object motion that the hanging tray
avoids.

\subsubsection{Liquid Handling}

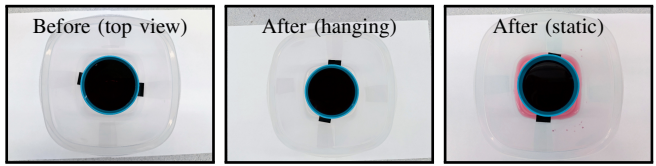
\begin{figure}[t]
  \vspace{3pt}
  \centering
  \footnotesize
  \begin{tikzpicture}
    \draw (-2, 0) node[inner sep=0] {\fbox{\includegraphics[height=0.8in,trim={0 0 0 0},clip]{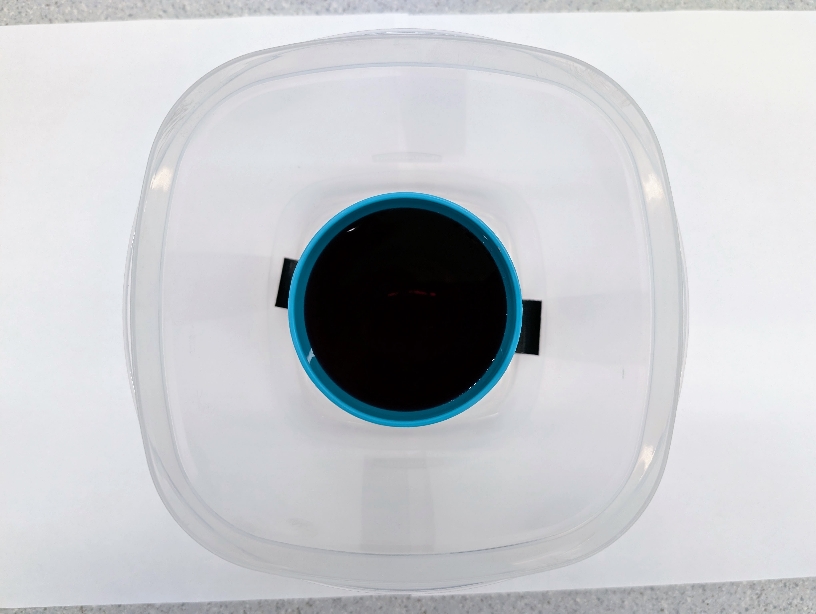}}};
    \draw (0.9, 0) node[inner sep=0] {\fbox{\includegraphics[height=0.8in,trim={0 0 0 0},clip]{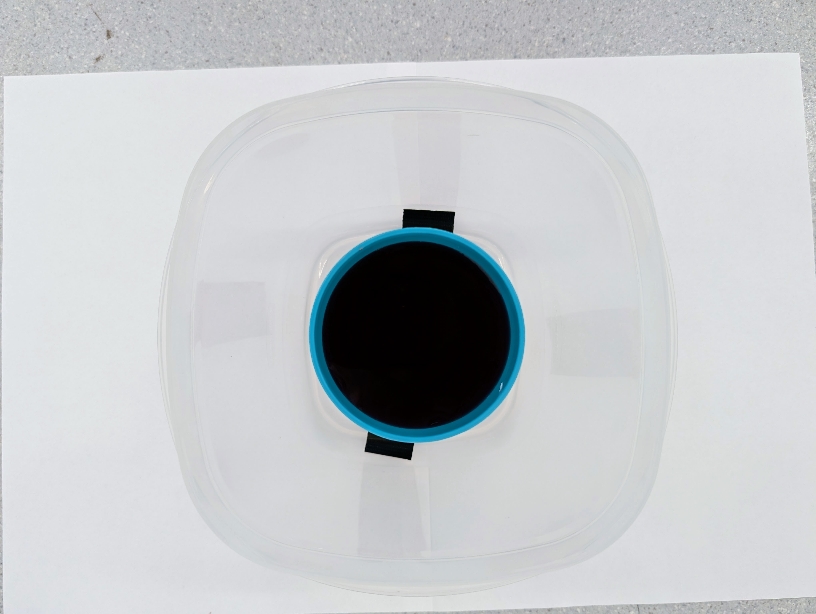}}};
    \draw (3.8, 0) node[inner sep=0] {\fbox{\includegraphics[height=0.8in,trim={0 0 0 0},clip]{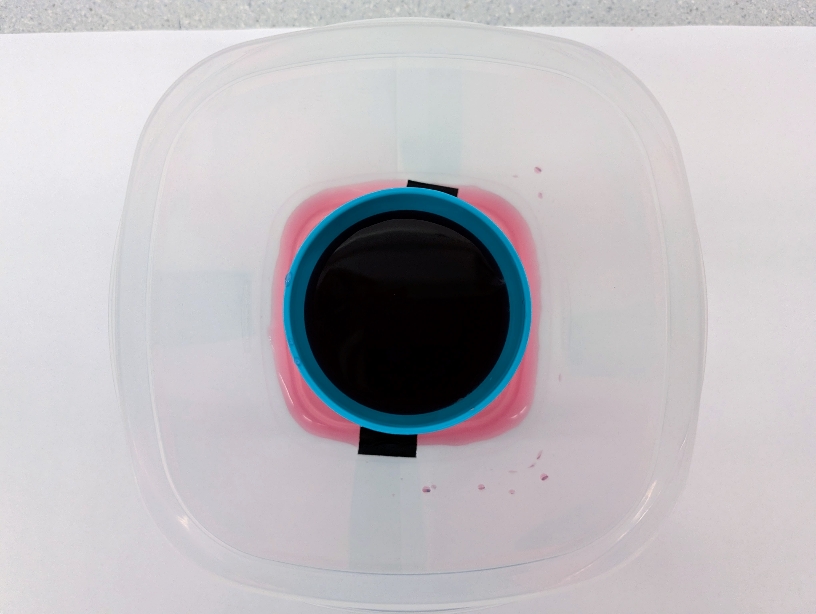}}};
    \draw (-2, 0.75) node {Before (top view)};
    \draw (0.9, 0.75) node {After (hanging)};
    \draw (3.8, 0.75) node {After (static)};
  \end{tikzpicture}
  \caption{Liquid handling experiments. The blue cup is rigidly
    set in the center of the larger plastic container and filled
    with~\SI{200}{ml} of water colored red with food dye. The water is free to
    slosh out of the cup but will be caught in the larger container. The
    container is rigidly attached to the tray during experiments. \emph{From left:}
    Top view before experiments; after executing the fast trajectory with
    the hanging tray; after executing it with the static tray. No liquid
    spilled from the cup when using the hanging tray, but a considerable amount
    was spilled with the static tray (the pink spill can be seen collected in the
    bottom of the larger container).}
  \label{fig:liquid_handling}
\end{figure}

Previous work~\cite{muchacho2022a} on pendular motion was focused on
transporting liquids while minimizing slosh and spillage, so we also perform an
experiment to assess the hanging tray's ability to transport liquid without
spilling. To avoid uncontrolled spills, we use a cup rigidly set within a
larger container that is in turn rigidly attached to the tray (see
Fig.~\ref{fig:liquid_handling}). For each experiment, the cup is filled
with~\SI{200}{ml} of water colored with red food dye, which is free to slosh
out of the cup (but any spill is captured inside the larger container). As
shown in Fig.~\ref{fig:liquid_handling}, the hanging tray results in no spill,
while a considerable amount of liquid sloshed out of the cup with the static
tray.

\subsection{Interactive Robot Waiter}

\begin{figure}[t]
  \centering
  \begin{tikzpicture}
    \draw (0, 0) node[inner sep=0] {\fbox{\includegraphics[height=1.2in,trim={0 0 0 0},clip]{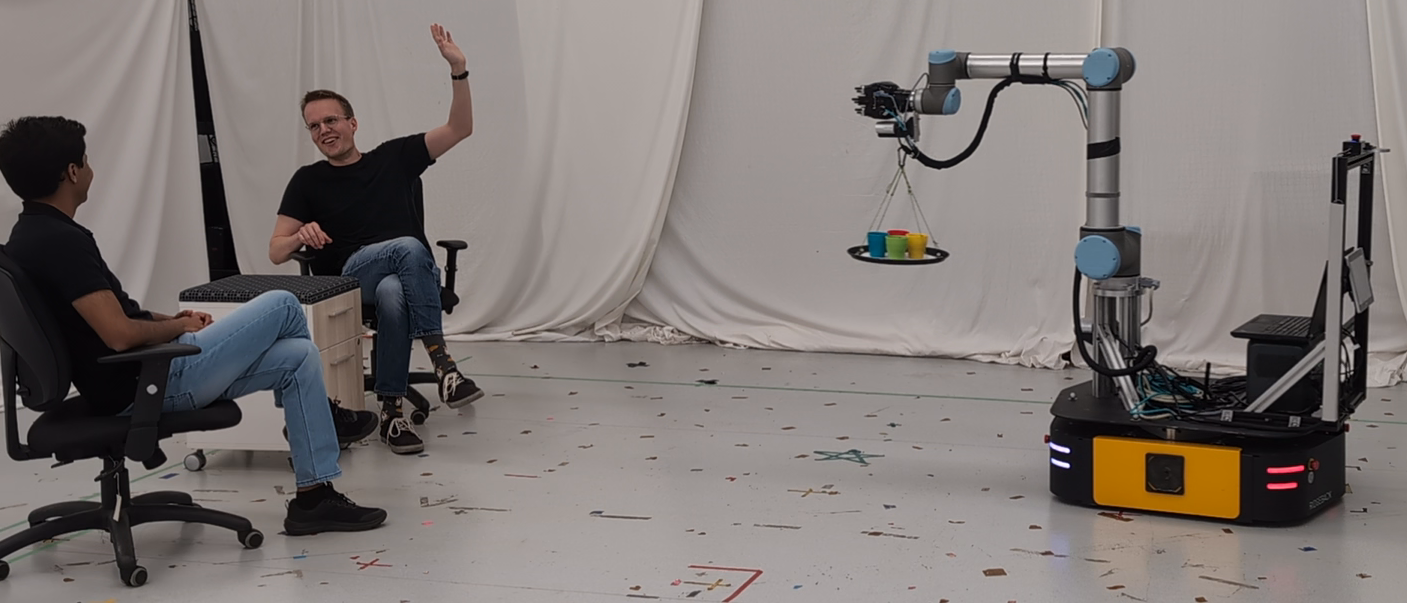}}};
    \draw (-1.74, -3.36) node[inner sep=0] {\fbox{\includegraphics[height=1.3in,trim={0 0 0 0},clip]{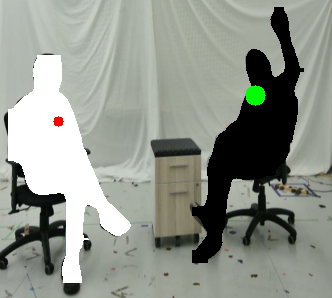}}};
    \draw (1.92, -3.36) node[inner sep=0] {\fbox{\includegraphics[height=1.3in,trim={0 0 0 0},clip]{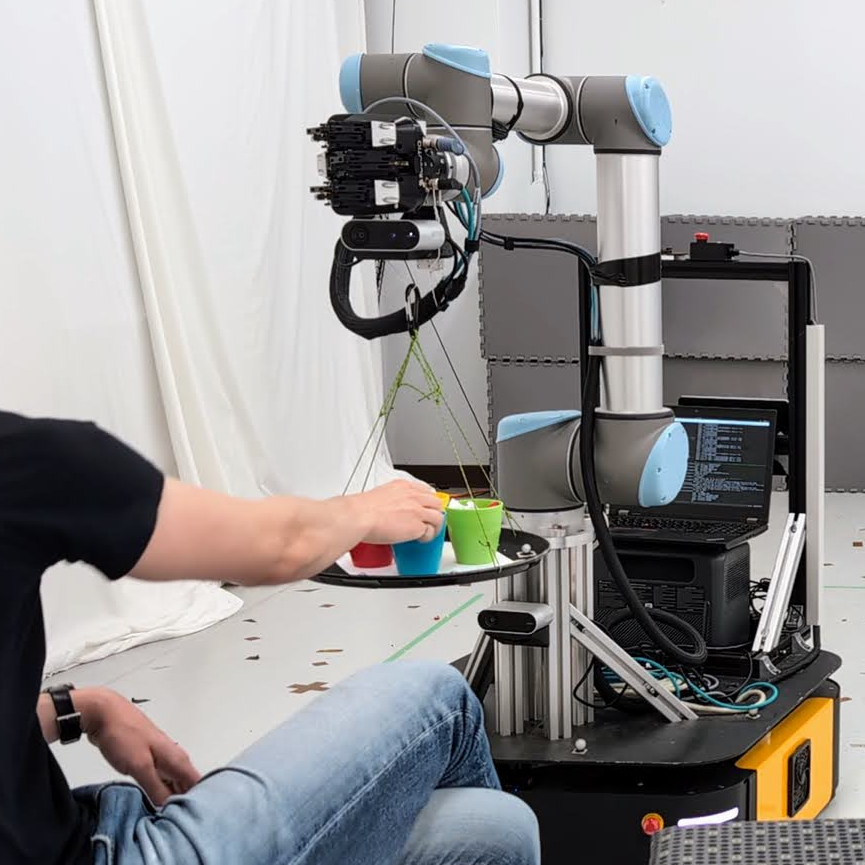}}};
  \end{tikzpicture}
  \caption{\emph{Top:} A person is raising his hand to indicate that he wants to
  be served by the robot and the robot is moving toward him.
  \emph{Bottom left:} The view from the robot's
  camera: the person with his hand up is segmented in black; the person with
  his hand down is segmented in white. The large green dot is the current
  target to steer toward using visual servoing. \emph{Bottom right:} The robot
  has reached the person and waits while a cup is taken from the tray. The full
  demonstration can be seen at \textbf{\texttt{\scriptsize \videourl}}.}
  \label{fig:demo}
  \vspace{-5pt}
\end{figure}

As a final experiment, we present a demonstration of our full interactive
waiter using visual servoing, as described in Sec.~\ref{sec:interactive}, with
parameters~$\bar{v}=\SI{0.3}{m/s}$ and~$k_\omega=0.5$. The robot is equipped
with the hanging tray carrying four cups while two people are sitting nearby.
An Orbbec Femto Bolt RGB-D camera attached to the EE is used to observe the
scene. Each person raises their hand in turn to summon the robot and take a cup
from the tray without leaving their seats. Some salient still images of the
experiment are shown in Fig.~\ref{fig:demo}, but we encourage the reader to
watch the accompanying video to see the full demonstration in action.

\section{Conclusion}

We presented an approach for robotic nonprehensile object transportation using
a hanging tray, the dynamics of which naturally reduce the shear forces acting
on the transported objects and prevent them from sliding, without requiring an
actuated robot arm. We demonstrated this approach using rigid objects, liquid,
and in an interactive robot waiter experiment.

\bibliographystyle{IEEEtran}
\bibliography{bibliography}

\end{document}